\documentclass[runningheads]{llncs}
\usepackage{eccv}
\usepackage{eccvabbrv}
\usepackage{graphicx}
\usepackage{booktabs}
\usepackage{multirow}
\usepackage[accsupp]{axessibility}
\usepackage{hyperref}
\usepackage{orcidlink}
\makeatletter
\newcommand{\printfnsymbol}[1]{%
  \textsuperscript{\@fnsymbol{#1}}%
}
\makeatother

\begin{document} 
\title{InstructGIE: Towards Generalizable Image Editing} 

\author{Zichong Meng\thanks{Equal Contribution}\inst{1} \and
Changdi Yang\printfnsymbol{1}\inst{1} \and
Jun Liu\inst{1} \and
Hao Tang\thanks{Corresponding Authors}\inst{2} \and
Pu Zhao\printfnsymbol{2}\inst{1} \and 
Yanzhi Wang\printfnsymbol{2}\inst{1}}

\authorrunning{Z.~Meng and C.~Yang et al.}

\institute{Northeastern University, Boston MA 02115, USA \and
Peking University, Beijing 100871, China \& \\ Carnegie Mellon University, Pittsburgh PA 15213, USA}

\maketitle
\begin{figure}
\centering
  \includegraphics[width=320 pt]{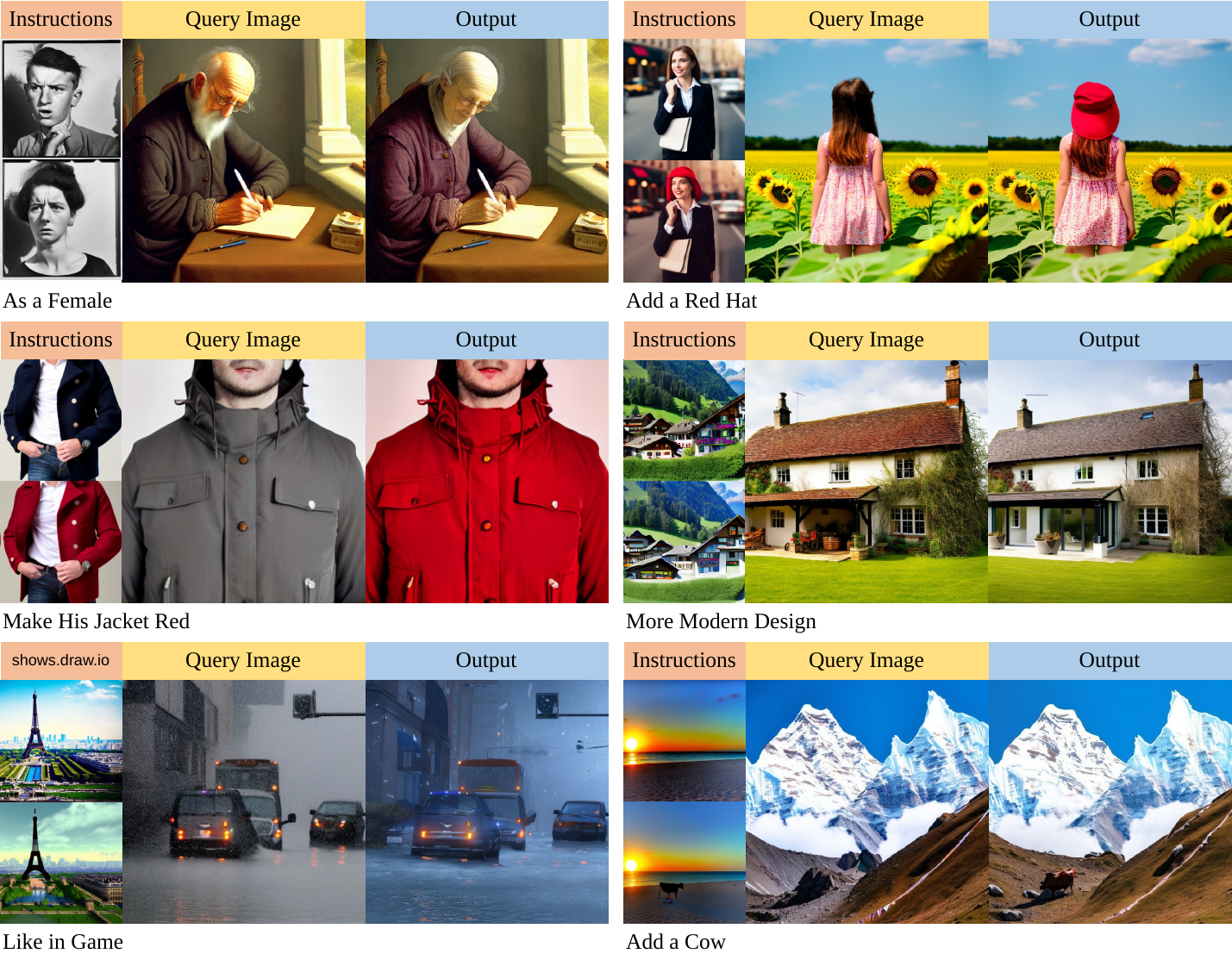}
    \caption{Demo results of the proposed\textbf{ InstructGIE} framework on various image manipulation tasks to both humans and scenes. By our proposed method, our model can generalize to generate the desired output with great detail qualities.}
    \label{fig:results_shows}
\end{figure}

\begin{abstract}
Recent advances in image editing have been driven by the development of denoising diffusion models, marking a significant leap forward in this field. Despite these advances, the generalization capabilities of recent image editing approaches remain constrained. In response to this challenge, our study introduces a novel image editing framework with enhanced generalization robustness by boosting in-context learning capability and unifying language instruction. 
This framework incorporates a module specifically optimized for image editing tasks, leveraging the VMamba block and an editing-shift matching strategy to augment in-context learning. Furthermore, we unveil a selective area-matching technique specifically engineered to address and rectify corrupted details in generated images, such as human facial features, to further improve the quality. Another key innovation of our approach is the integration of a language unification technique, which aligns language embeddings with editing semantics to elevate the quality of image editing.
Moreover, we compile the first dataset for image editing with visual prompts and editing instructions that could be used to enhance in-context capability.
Trained on this dataset, our methodology not only achieves superior synthesis quality for trained tasks, but also demonstrates robust generalization capability across unseen vision tasks through tailored prompts. Our project page is available at \url{https://cr8br0ze.github.io/InstructGIE}.

\end{abstract}
\section{Introduction}
As a crucial task in computer vision, image editing has witnessed significant improvements enhanced with the increasingly popular denoising stable diffusion techniques in recent years\cite{imagebrush, painter, promptdiffusion,liu2023more, kawar2023imagic}. Given a set of text or image prompts as generation constraints or instructions,  diffusion-based image editing can follow the instructions and synthesize an edited image.  However, since the model does not have the capability to accurately model all possible samples in the conditional distribution space~\cite{survey}, if specific instructions are not included in the training dataset,  current diffusion-based image editing methods can hardly generate satisfactory results. Thus, editing performance largely depends on the training dataset without superior generalization capabilities.    

On the other hand, large language models (LLMs) have proven extraordinary abilities to   learn  from contexts,  referred to as in-context learning, which allows LLMs to perform unseen tasks by providing a combination of input-output examples and a query input.  Inspired by the potential to enhance the generalization of the model with LLMs, \cite{imagebrush, promptdiffusion} explore in-context learning for computer vision tasks, allowing them to perform unseen tasks with novel vision-language prompt designs. However, these methods are not tailored for image editing applications, leading to unsatisfying synthetic qualities with inaccurate or incorrect output and lack of detail.   
To improve the generalization of image editing with improved synthetic image quality, it is crucial to effectively understand the text \& image prompts and specifically control image editing details, which is challenging in the current literature.   

In this work, we propose InstructGIE, an image editing framework with enhanced generalizability. We improve image editing performance from both visual and text aspects. (i) For the visual information, we incorporate a VMamba-based module to specifically enhance the image editing outputs. As VMamba\cite{vmamba} has proven its better performance in capturing in-context information from inputs with larger  receptive fields\cite{vmamba},  we leverage VMamba  and  propose an editing-shift matching strategy to augment in-context learning. Furthermore, since current image editing works do not perform well in generating correct features with accurate details,  we unveil a selective area-matching technique specifically engineered to address and rectify corrupted details in generated images, such as human facial features, to further improve the quality. 
(ii) Another key innovation of our approach is the integration of a language unification technique,  which aligns language embeddings with editing semantics to elevate the quality of image editing. Our framework not only achieves superior in-context generation for trained tasks but also demonstrates robust generalization across unseen vision tasks. Moreover, we compile a publicly available image editing dataset with plenty of visual prompts and editing instructions for better generalization evaluation of image editing. Our contributions are summarized as follows:

\begin{itemize}
    \item We propose an image editing framework, including in-context learning enhancement and language unification strategies, specifically designed to enhance generalization ability from both visual and text domains. 
    \item We compile the first dataset for image editing with visual prompts and editing instructions that could be used to enhance in-context capability.
    \item We conduct extensive experiments and achieve great generalization ability in the multiple unseen image editing task, both quantitatively and qualitatively.
\end{itemize}
\label{sec:introduction}
\section{Related Works}
\subsection{Denoising Stable Diffusion Based Image Editing}
Denoising Stable Diffusion\cite{ho2020denoising, sohl2015deep, song2019generative} based image editing could follow guidance from text or image prompts. With the foundation of text-guided models offering rich generative capabilities, there has been a surge in research aimed at adapting these models for image manipulation tasks from textual descriptions. To steer the image editing process in the desired direction, the use of models like CLIP to fine-tune diffusion models has become a common practice. Although these methods\cite{brooks2023instructpix2pix, promptdiffusion, kawar2023imagic} have shown impressive results, they often involve costly fine-tuning processes. Recent innovations\cite{hertz2022prompt} have introduced techniques that inject cross-attention into the models to more effectively edit specific semantic areas within the spatial feature maps. Further advancements\cite{kwon2022diffusion} have enhanced these techniques by adding semantic loss or applying attention loss to refine the integration of plugged features, improving the precision and quality of the editing outcomes. \cite{imagebrush} proposes a framework that could learn instructions from visual image pairs for more accurate editing and firstly formulate this task as an image inpainting problem. 

\subsection{Generalization Capability in Visual Tasks}
In-context learning is widely applied in the field of natural language processing (NLP), enabling models to adapt to new tasks such as translation, question answering, and complex reasoning. NLP models utilize in-context examples, comprising text and corresponding labels, to tackle tasks they haven't seen before. However, applying in-context learning to the visual domain introduces more challenges and remains less explored. 
A significant hurdle is the nature of fixed-size input requirements for vision models, as opposed to variable-length text inputs that can be managed by language models. Vision models generally struggle with processing inputs of varying sizes, making it impractical to process multiple image prompts in one-shot for global understanding.
Moreover, in intricate visual understanding, specific instructions are often implied from a limited set of image examples rather than explicitly stated, which poses additional difficulties for vision models in identifying and understanding high-level visual relationships. 

Recent strides in applying masked image modeling have marked a step forward in improving in-context learning for vision models. The method proposed by \cite{promptdiffusion}, employing a masked autoencoder-based technique, predicts a missing image within a two-by-two grid, using two images as in-context examples and another as the query. This concept was later expanded by \cite{imagebrush} with a multitask framework. Despite their progress, such inpainting methods are limited by the necessity of a fixed number of in-context examples and increased memory demands. Painter, highlighted in \cite{painter}, exemplifies an inpainting approach tailored for versatility across various vision tasks.

In contrast, inspired by ControlNet \cite{controlnet}, \cite{promptdiffusion}  refines the  framework by adding an additional pair of example images and employing a multitask supervised finetuning method.
Prompt diffusion excels in visual in-context learning. 
However, it faces certain limitations or challenges in its practical applications.

\subsection{Dataset for Diffusion-based Image Editing}
Currently, various types of datasets are used for training in diffusion-based image editing. There are datasets that concentrate on specific domains like CelebA~\cite{liu2015faceattributes} and FFHQ~\cite{ffhq} for human face image manipulation, AFHQ~\cite{afhp} for animal face image editing, LSUN~\cite{lsun} for object modification, and WikiArt~\cite{mohammad2018wikiart}   for style transfer. In-the-wild video datasets could also be leveraged to train image editing tasks. The Scannet dataset~\cite{dai2017scannet} encompasses a vast array of more than 1,500 indoor scenes from various settings, such as apartments, offices, and hotels, providing extensive annotations. The LRW dataset~\cite{chung2017lip}, tailored for lip reading tasks, includes more than 1000 video utterances of 500 distinct words. The UBC-Fashion dataset \cite{zablotskaia2019dwnet} features 600 videos spanning various clothing categories, with 500 videos allocated for training and 100 for testing, guaranteeing no repetition of individuals in the training set. The DAVIS dataset\cite{zablotskaia2019dwnet} (Densely Annotated VIdeo Segmentation), a widely recognized benchmark for video object segmentation, contains 150 videos in total. There are also image editing works proposing to generate datasets with editing instructions. InstructPix2pix~\cite{brooks2023instructpix2pix} collects over 450,000 training image pairs. For each pair,  given an image with its caption, it first uses a finetuned GPT-3~\cite{gpt3} to generate an editing instruction and an edited image caption. Then it employs Stable Diffusion and the Prompt-to-Prompt algorithm~\cite{hertz2022prompt} to generate edited image  following the caption.
However, currently there are no datasets with multiple image pairs under one editing instruction, which is crucial to enhance the generalization ability of image editing. 
\label{sec:related work}
\section{Preliminary}
Recent advances in generative models have been significantly driven by the emergence of diffusion models, which have set new benchmarks in image creation\cite{dhariwal2021diffusion, kawar2022enhancing, song2020score}. These models have found applications across a broad spectrum of areas\cite{chen2022re, zimmermann2021score, theis2022lossy, blau2022threat, kawar2022denoising}, demonstrating their versatility and effectiveness. The fundamental concept behind diffusion models involves starting with an image that is initially just random noise $\mathbf{x}_T \sim \mathcal{N}(0, \mathbf{I})$ and progressively refining this image step by step until it becomes a high-quality, realistic image $\mathbf{x}_0$. This refinement process involves generating intermediate samples $\mathbf{x}_t$ (for $t \in\{0, \ldots, T\}$), where each sample is defined as:
\begin{equation}
\mathbf{x}_t=\sqrt{\alpha_t} \mathbf{x}_0+\sqrt{1-\alpha_t} \boldsymbol{\epsilon}_t,
\end{equation}
where the parameter $\alpha_t$  sets the pace of the diffusion process, ranging from $0=\alpha_T<\alpha_{T-1}<\cdots<\alpha_1<\alpha_0=1$, and $\boldsymbol{\epsilon}_t \sim \mathcal{N}(0, \mathbf{I})$ represents the added noise. The model refines the image by applying a neural network $f_\theta\left(\mathbf{x}_t, t\right)$ to each sample $\mathbf{x}_t$, followed by the addition of Gaussian noise to produce the next sample $\mathbf{x}_{t-1}$. This neural network is optimized to achieve a denoising goal, striving for $f_\theta\left(\mathbf{x}_t, t\right) \approx \epsilon_t$, resulting in a generative process that closely mimics the desired image distribution.

Expanding this framework to conditional generative modeling, the process involves conditioning the neural network $f_\theta\left(\mathbf{x}_t, t, \mathbf{y}\right)$ on an additional input $\mathbf{y}$, enabling the generation of images from a distribution conditioned on $\mathbf{y}$. This conditional input could be anything from a low-resolution image, a category label, or a descriptive text sequence. Leveraging the advancements in LLMs \cite{raffel2020exploring} and hybrid vision-language models \cite{radford2021learning}, text-to-image diffusion models are developed. These models allow for the creation of detailed, high-resolution images from mere text descriptions, starting with a low-resolution image generated through the diffusion process, which is subsequently refined into a high-resolution image using additional models.
\label{Preliminary}
\section{The Proposed Method}
\label{sec:methods}
\begin{figure*}[t]
\centering
  \includegraphics[width=\textwidth]{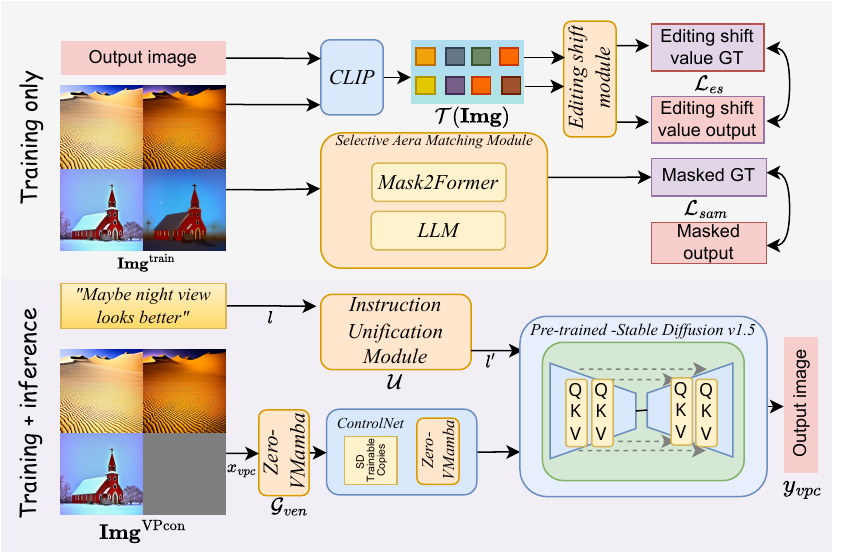}
    \caption{Overall architecture of InstructGIE. The lower pipeline is for both training and inference processes where the model obtains unified editing instructions outputted by  Instruction Unification Module $\mathcal{U}$ and combines with visual prompted input $\textbf{Img}^{\text{VPcon}}$ to pass through Zero-VMamba integrated Stable Diffusion model with ControlNet for output image. The upper pipeline is for training only which compares output image and training ground truth $\textbf{Img}^{\text{train}}$ and computes editing shift loss $\mathcal{L}_{es}$ with Editing Shift Module and selective area matching loss $\mathcal{L}_{sam}$ with Selective Area Matching Module.}
    \label{fig:architecture}
\end{figure*}

We present our framework pipeline in \cref{fig:architecture}. For efficient training and better controllability,  we adopt a line of techniques with popular architectures such as ControlNet~\cite{controlnet} and Stable Diffusion~\cite{sd} to design a generalizable image editing tool with accurate high-quality outputs. Specifically,  we introduce enhanced in-context learning both at the architecture level and training level
to improve the image quality. Furthermore, language instruction unification is adopted to maximize the generalization ability for unseen editing tasks.  Moreover, selective area matching is proposed to further improve the output quality with full details.

\subsection{Enhanced In-Context Learning}
\label{sec:eil}
Visual prompting based on inpainting is an effective visual in-context learning method   in various computer vision tasks~\cite{vqgan-mae, painter}, which is applied    in image editing tasks~\cite{imagebrush} recently. However, the methods perform poorly in quality when dealing with unseen image manipulation tasks. Therefore, we propose the enhanced in-context learning specifically tailored  for generalizable image editing.
\begin{figure*}[t]
\centering
  \includegraphics[width=300 pt]{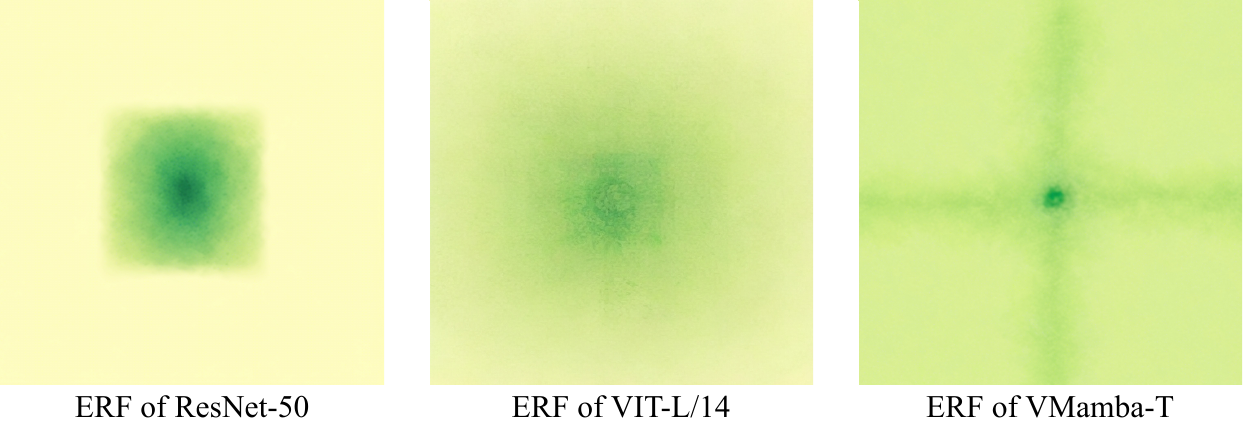}
    \caption{\textbf{Effective Reception Field (ERF)} of ConvNet, ViT, VMamba based model architectures. }
    \label{fig:erf}
    % \vspace {-0.5cm}  
\end{figure*}
\subsubsection{Reforming Conditioned Latent Diffusion Model.}
To improve the generalization ability of image editing, it is crucial for the framework to  explicitly capture low-level visual editing contexts. Current diffusion-based image editing methods~\cite{imagebrush, visii} that involve visual prompting either adopt ConvNet~\cite{convnet} or ViT~\cite{vit} as the vision encoder for visual prompts. However, these methods fail to generalize well as they are not able to capture enough visual editing contexts (see Fig.~\ref{fig:erf}). To address this, we formulate the visual  prompted condition as a single image $\textbf{Img}^{\text{VPcon}}$ = \{$\textbf{Img}_{0}^{\text{in}},$ $\textbf{Img}_{0}^{\text{out}},$ $\textbf{Img}_{1}^{\text{in}},$ $\textbf{Grey}$\},  as shown in Fig~\ref{fig:architecture}, with a global effective receptive field (ERF).

Moreover, we further propose to reform the conditioned latent diffusion model. Inspired by the recent visual state space model VMamba~\cite{vmamba} which exhibits a better global ERF and also emphasizes shifting boundaries of input’s four quadrants, we propose to adopt a vision encoder based on Zero-VMamba to fit our structure. Specifically, the vision encoder $\mathcal{G}_{ven}$   comprehends   the visual prompted condition in latent space $x_{vpc}$ as follows, 
\begin{equation}
    e_{vpc} = \mathcal{G}_{ven}(x_{vpc};\Theta_{g}),
\end{equation}
where $e_{vpc}$ is the processed embedding of the  visual prompted condition, and $\Theta_{g}$ is the model parameters   initialized to zeros.

To further improve the performance, after each ControlNet trainable copied modules $\mathcal{F}_{c}$ with parameters $\Theta_{c}$, we  propose to link and inject the processed visual prompted condition information to the frozen Stable Diffusion model $\mathcal{F}$ with parameters $\Theta$ through zero-VMamba layer $\mathcal{G}$. 
We use two instances of VMamba with parameters $\theta_{g1}$ and $\theta_{g2}$
respectively. The complete model then computes the following,
\begin{equation}
    y_{vpc} = \mathcal{F}(x;\Theta) +  \mathcal{G}(\mathcal{F}(x+\mathcal{G}(x_{vpc};\Theta_{g1});\Theta_{c});\Theta_{g2}),
\end{equation}
where $y_{vpc}$ is the output of our conditioned diffusion model block. 

Our conditioned latent diffusion model can  process all four quadrants in our visual prompted conditions with a global receptive field, while it does not generate  random noises during initial training steps. 

\subsubsection{Editing-Shift Matching.}
\label{esm}
Besides the architecture innovation, we  incorporate an editing-shift-matching technique to enhance in-context learning ability in image editing with more accurate detailed outputs.

In specific, for each training ground truth $\textbf{Img}^{\text{train}}$ = \{$\textbf{Img}_{0}^{\text{in}},$ $\textbf{Img}_{0}^{\text{out}},$ $\textbf{Img}_{1}^{\text{in}},$ $\textbf{Img}_{1}^{\text{out}}$\}, we calculate a implicit editing shift value using CLIP~\cite{clip} image embedding:
\begin{equation}
    \mathcal{T}(\textbf{Img}) = \frac{\sum_{i = 0}^1 \text{CLIP}(\textbf{Img}_{i}^{\text{in}}) - \text{CLIP}(\textbf{Img}_{i}^{\text{out}})}{2}.
\end{equation}
During the training process, after predicting the noise, we use it to reverse the noised input and obtain a pseudo output image $\textbf{Img}^{\text{PO}}$ = \{$\textbf{\text{PO}}_{0}^{\text{in}},$ $\textbf{\text{PO}}_{0}^{\text{out}},$ $\textbf{\text{PO}}_{1}^{\text{in}},$ $\textbf{\text{PO}}_{1}^{\text{out}}$\}. Our framework then calculates the editing transfer value of the pseudo output image and deduces a editing shift loss to optimize during our training  via the cosine similarity of the two design transfer values:
\begin{equation}
    \mathcal{L}_{es} = 1- \frac{\mathcal{T}(\textbf{Img}^{\text{PO}}) \cdot \mathcal{T}(\textbf{Img}^{\text{train}})}{||\mathcal{T}(\textbf{Img}^{\text{PO}})||\times||\mathcal{T}(\textbf{Img}^{\text{train}})||}
\end{equation}

Through editing-shift matching, our model can better comprehend how editing should be done within a visual context level through an implicit editing shift value. Furthermore, this implicit editing shift value can further guide the sampling process creating a controllable editing.

\subsection{Language Instruction Unification}
\label{sec:liu}
Previous works that utilize visual prompting in image editing tend to focus more on visual instructions and believe language instructions can be vague and unstable~\cite{imagebrush}. 
We believe that the potential of text prompts is not fully explored in image editing.
Language instructions  have  significant impacts on diffusion model outputs.   Language instructions with the same meaning can still result in entirely different outputs due to different processed language embeddings.

To improve image editing and explore language instructions,  we propose a novel approach, language instruction unification. 
During the training process, for each batch of training data, we randomly sample  50\% of training data, collect their language editing instructions $l$, and process them through a frozen lightweight  LLM, Open Llama 3b V2 Quant 4~\cite{openllama} denoted as $\mathcal{U}$. The LLM $\mathcal{U}$ is fixed prompted with a fixed random seed   to uniformly reformulate the language instruction $l$ better for machine-level understanding. The LLM $\mathcal{U}$ will output a unified language editing instruction $l^\prime$.
\begin{equation}
    l^\prime = \mathcal{U}(l)
\end{equation}

We then augment the training data with unified language editing instructions.

During the inference, each language editing instruction $l$  is  passed through the frozen LLM $\mathcal{U}$ for language instruction unification and then sent to  our conditioned diffusion model.

By adopting language instruction unification for training data augmentation during the training, our conditioned diffusion model can  learn diverse non-uniformed editing instructions to build up the model's knowledge distribution with unified language prompts.
Adopting language instruction unification during inference aligns with the training, therefore, greatly minimizing the possibility of diverse quality in edited outputs and maximizing the ability to generalize to unseen editing tasks.

\subsection{Selective Area Matching}
Diffusion-based image editing models usually suffer from low quality in specific details and this bottleneck appears to be more crucial in generalizable image editing methods~\cite{imagebrush, visii}. The details of  human and animal images are easily distorted in these methods. 

A naive solution  might be utilizing  negative prompts~\cite{sd} like `do not draw hands with 6 fingers' in general text-to-image tasks. However, in image editing,  it is challenging to apply negative prompts. Users typically can not foresee the specific details after-edited outputs, therefore they are not able to construct appropriate  negative prompts. Besides,  negative prompts may   limit the artistic potential of image editing models.

To properly address this issue for generalizable image editings, we propose an  optimization method, namely selective area matching, that targets the difference in the chosen detailed editing area between the original training ground truth $\textbf{Img}^{\text{train}}$ and the reversed pseudo output $\textbf{Img}^{\text{PO}}$.

In particular, during the training process, we incorporate a frozen Mask2Former model\cite{mask2former} $\mathcal{M}$ to obtain panoptic segmented training image information including segmented masks and class information $C$.
\begin{equation}
    \text{segmask}, C = \mathcal{M}(\textbf{Img}^{\text{train}})
\end{equation}
After that, our framework  processes the class information using the same lightweight LLM $\mathcal{U}$ described in   Sec.~\ref{sec:liu} to filter out pre-defined classes  including living creatures and  humans requiring special attention for addressing the details.
\begin{equation}
C_{\text{filtered}} = \{ c \in C | \mathcal{U}(c, \text{selected classes}) = 1 \}
\end{equation}
Based on selected class labels, the framework   then deduces a segmented binary mask for the selected editing area.
\begin{equation}
[\text{mask}]_{i} =
\begin{cases}
1 & \text{if } cls([\text{segmask}]_{i}) \in C_{\text{filtered}} \\ 0 & \text{otherwise}
\end{cases}
\end{equation}
During the training process, our framework calculates and optimizes the selective-area matching loss by
\begin{equation}
    \mathcal{L}_{sam} = \frac{1}{N} \sum_{i=1}^{N} (([\textbf{Img}^{\text{PO}}]_i \cdot \text{[mask]}_i) - ([\textbf{Img}^{\text{train}}]_i \cdot \text{[mask]}_i))^2
\end{equation}
where $N$ as the total pixel number in the image.

With selective area matching, image editing does not need negative prompts to deal with distorted details in images, which can make the most of the model's artistic editing capacity to generate high-quality outputs with great details. It is only  incorporated during training, which does not change the inference, greatly saving inference efforts compared with negative prompts.

\section{Visual Prompt Dataset For Image Editing}
\begin{figure*}[t]
\centering
  \includegraphics[width=\textwidth]{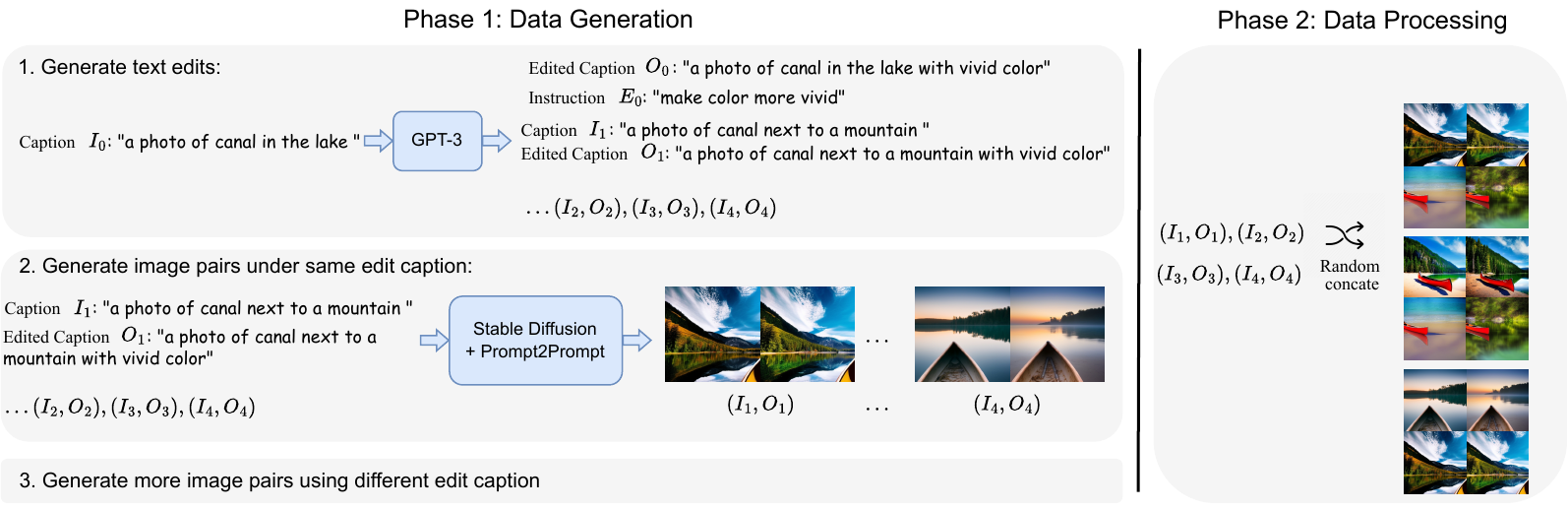}
    \caption{\textbf{Dataset generation process} Our dataset generation consists of two phases. Data Generation is to generate sets of image pairs under one editing caption. Data Processing is randomly pick image pairs under the same editing instruction and concatenate them together as one input for training. }
    \label{fig:data_generation}
\end{figure*}

Traditional image editing datasets only contain editing image pairs with a small amount of similar editing instructions. To the best of our knowledge, there is no open-sourced image editing dataset that is explicitly designed for image editing with visual prompting, which utilizes multiple different image pairs for each editing instruction to provide a general demonstration in various cases.

Therefore, we introduce and open source a new dataset that is designed specifically for image editing tasks utilizing visual prompts. Our dataset generation pipeline is shown in Fig.~\ref{fig:data_generation} with 2 phases: Data Generation and Data Processing. In Data Generation phase, we first fine-tune  GPT-3~\cite{gpt3} for 2 epochs with 700 groups of human-written edits, each consisting of 5 different pairs of input and edited captions with one editing instruction. Then as shown in step 1, we generate around 3500 groups of editing prompts using the fine-tuned GPT-3 model. In each group, there is one instruction $E_{0}$ and five pairs of caption and edited caption $(I_{0}, O_{0}), (I_{1}, O_{1}),..., (I_{4}, O_{4})$.  In step 2, similar to InstructPix2Pix, we then also adopt Prompt-to-Prompt for image generation. For each input and edited caption pair, we generated 50 times each with random noise and followed InstructPix2Pix to keep the best one image pair using CLIP-based metric. In addition, we also make sure for each editing instruction, there are at least two pairs of images. In step 3, we generate more image pair sets using in the same way. With filtering, we obtained around 12,000 images with around 2,000 editing instructions. In data preparing, we randomly choose 2 pairs of images under the same editing instruction and concatenate them for training.
\section{Experiments}
\label{sec:experiment}
\subsection{Experimental Settings}
\begin{figure*}[t]
\centering
  \includegraphics[width=\textwidth]{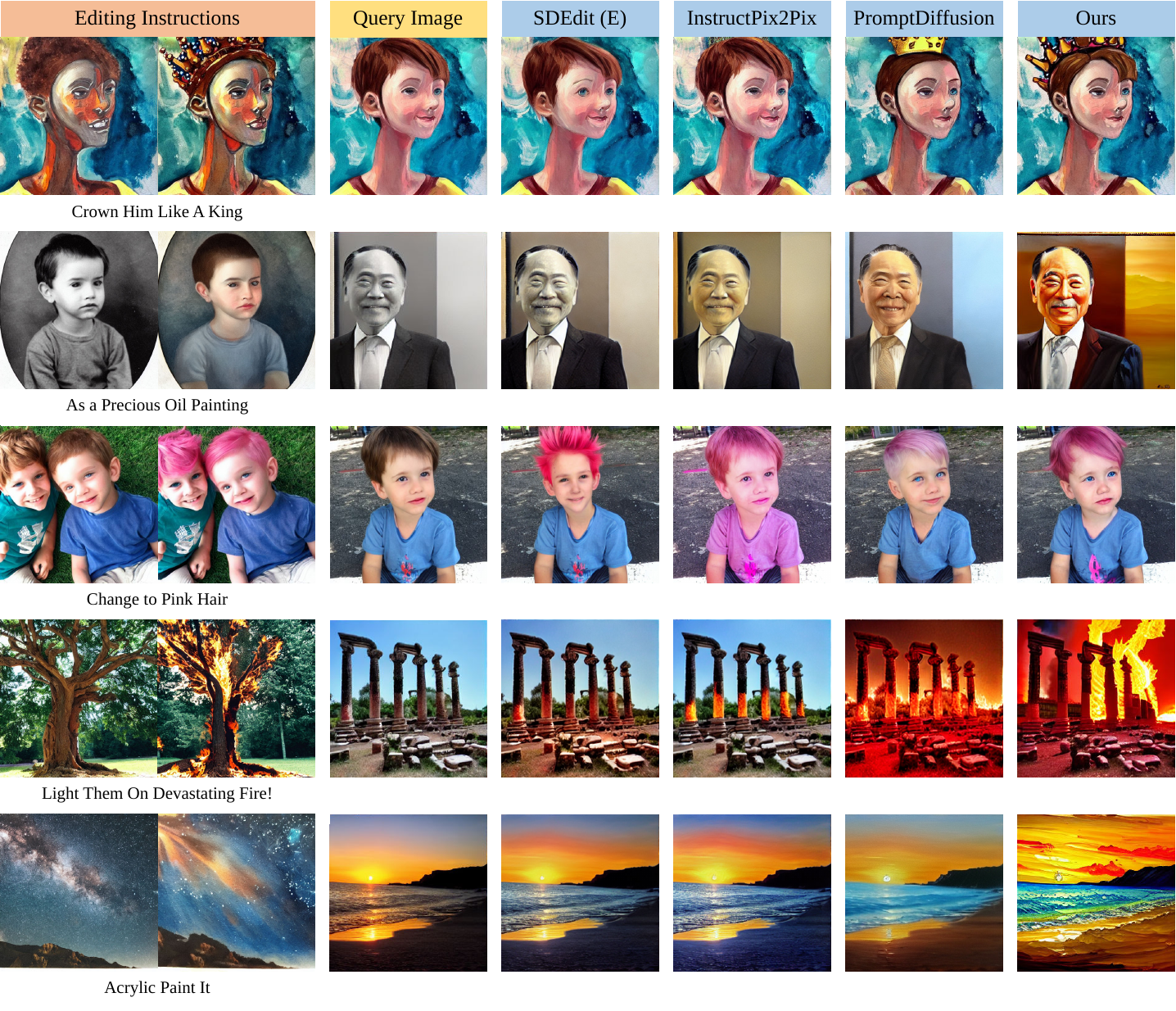}
    \caption{\textbf{Qualitative Comparison on our Test Dataset.} We conducted experiments on various scenarios, including human, architecture and landscape. }
    \label{fig:results_in_domain}
\end{figure*}
\textbf{Datasets.}
To fairly compare our methods with baseline methods, we conduct both qualitative and quantitative experiments  on our proposed synthetic dataset that involves over 12,000 filtered images and 2,000 filtered editing instructions. All single input images have a resolution of $512\times512$ and are resized to $256\times256$ for training and testing purposes.

\noindent\textbf{Implementation Details.}
In our approach, we split the dataset with 80\% for training  and  20\%  for testing. 
As demonstrated in \cref{fig:architecture}, two image pairs are concatenated into one $512\times512$ image with the same editing instructions as $\textbf{Img}^{\text{train}}$ and mask the fourth quadrant with a grey color as $\textbf{Img}^{\text{VPcon}}$.
We prepare the in-domain test dataset in the same format. For out-of-domain testings, we ensure that both   the  visual instruction pairs and the text instructions  are not used during training, to best simulate how models perform in real-life image editing generalization scenarios.

For baselines, we use their original configurations to train their model. In our method, we only fine-tune the additional ControlNet for 5000 steps with a batch size of 1024 and a learning rate at $1\times10^{-4}$. During
training, we set the classifier-free scale the same as the original ControlNet. And we randomly drop 15\% of language or visual editing instructions  to further enhance our model's generalization ability. Our implementation utilizes PyTorch and is trained on 4 Tesla A100-40G GPUs with AdamW optimizer.

\noindent\textbf{Comparison Methods.}
To evaluate the effectiveness of our work, we compare  with other state-of-the-art image editing frameworks, including SDEdit\cite{meng2021sdedit}, Instruct-Pix2pix\cite{brooks2023instructpix2pix} and PromptDiffusion\cite{promptdiffusion}. We adopt two quantitative metrics Fr’echet Inception Distance (FID) and CLIP directional Similarity (CLIP DirSim) proposed by Gal \etal~\cite{clipdirsim}. We utilize the FID score to quantitatively represent the image quality of   generated editing outputs, and CLIP DirSim  to evaluate how well the models follow editing instructions to produce the output.

\subsection{State-of-the-Art Comparisons}
\textbf{Qualitative Evaluation.}
In Fig.~\ref{fig:results_in_domain}, we present our qualitative results in the testing set (in domain). The comparison shows that our method surpass previous baseline methods. Our method  understands and follows both visual and language editing instructions better, and produces a far more detailed manipulated output especially in human figures.
\begin{figure*}[t]
\centering
  \includegraphics[width=\textwidth]{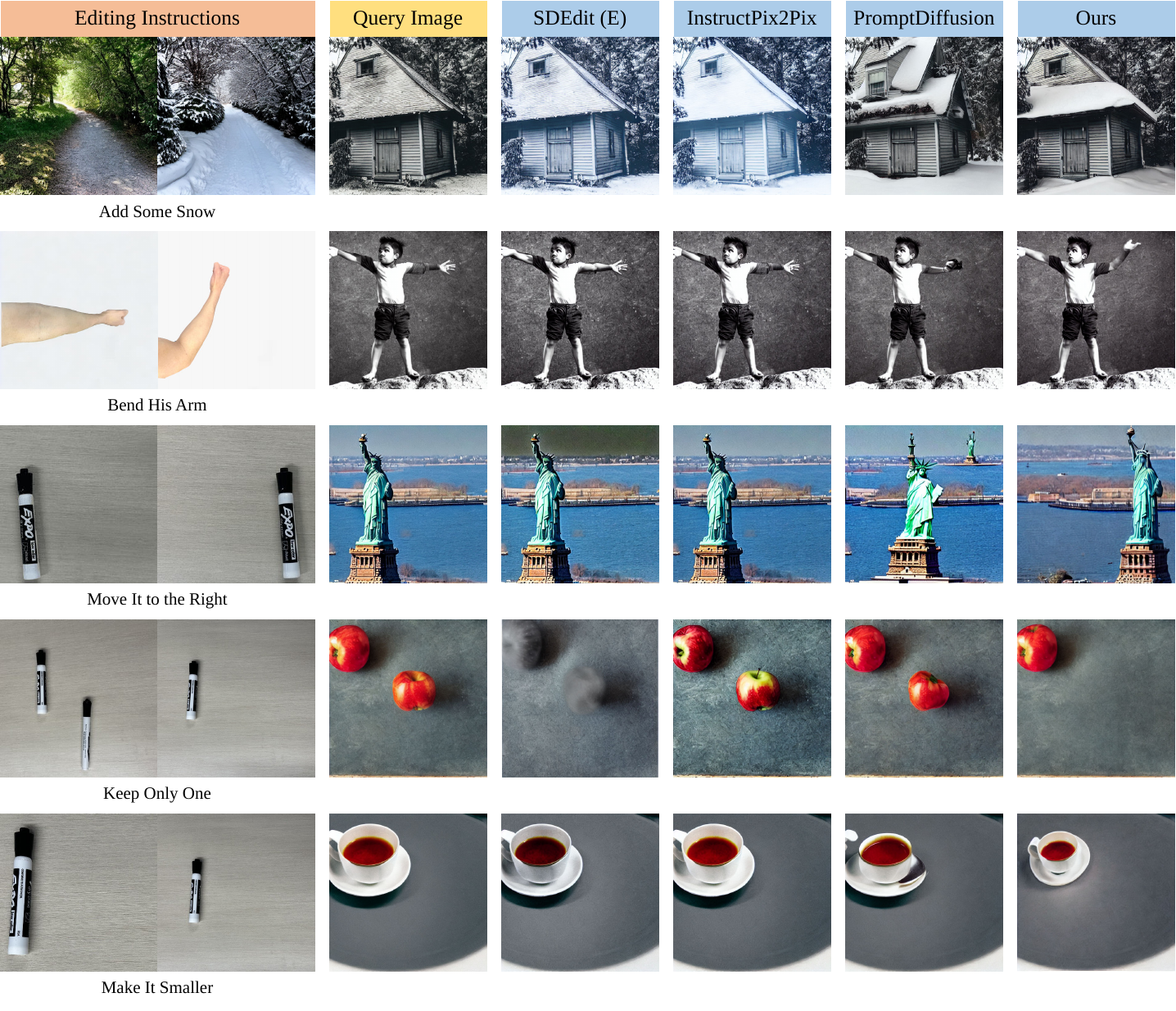}
    \caption{\textbf{Qualitative Comparison on Out-of-Domain Images.} We conducted experiments on images and instruct that are not in training dataset.}
    \label{fig:results_generalization} 
\end{figure*}
In Fig.~\ref{fig:results_generalization}, we present our qualitative results tested in out-of-domain settings. We include five editing instructions that are considered extremely hard for diffusion-based image editing model~\cite{sdsurvey}, including object add, object remove, structure location change, structure action change, and object size change. It is important to note that due to how we generate our training dataset utilizing Prompt-to-Prompt~\cite{prompt2prompt}, editing images pairs with these types of editing instructions is not feasible to generate   our training data. This generalization comparison shows that our method excels baseline methods by a significant margin. Our method shows a great capability to carry out well-detailed quality outputs following these hard editing instructions in diffusion-based image editing models, while other baseline methods all fail to understand the editing instructions well or perform manipulations close to the editing instructions.
\begin{figure*}[t]
\centering
  \includegraphics[width=\textwidth]{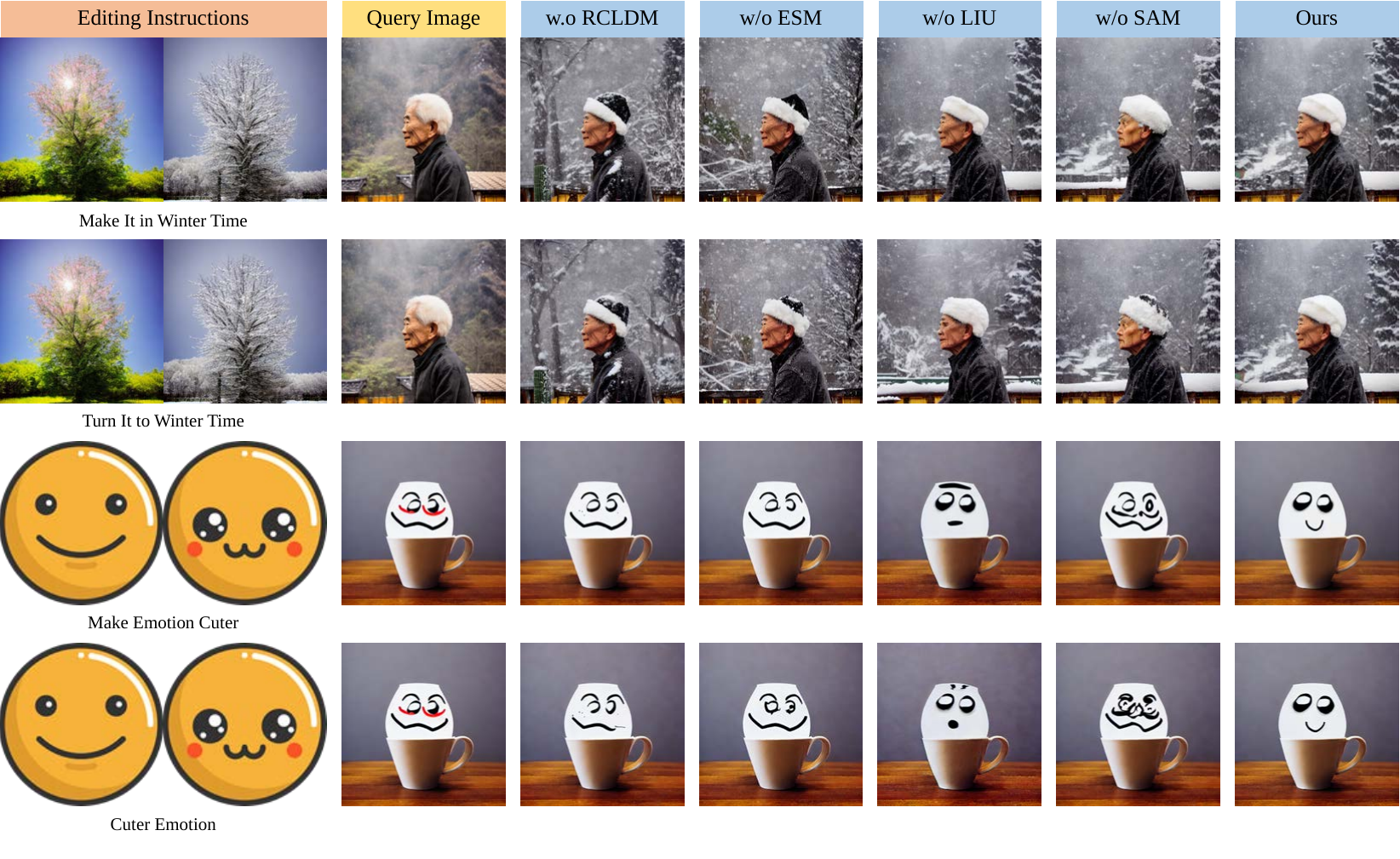}
    \caption{\textbf{Ablation Study Results} for both in-domain (first two rows), and out-of-domain first two rows) image manipulations}
    \label{fig:results_ablation} 
\end{figure*}

\noindent\textbf{Quantitative Evaluation.}
{\setlength{\tabcolsep}{8.5pt}
\renewcommand\arraystretch{1.25}
\begin{table}[tb]
  \centering
  \caption{Quantitative results comparison between our method and baseline methods. Quantitative results shows our method excels the baseline methods in both FID and CLIP directional Similarity score in a great margin.}
  \label{tab:test}
  \begin{tabular}{l|c|c|c|l}
  \hline
& SDEdit (E) & InstructPix2Pix & PromptDiffusion & Ours \\
\hline
FID $\downarrow$ & 21.67 & 17.87& 13.75  & 7.57 \\
\hline
CLIP DirSim $\uparrow$ & 0.11 & 0.17 & 0.21 & 0.27 \\
  \hline
  \end{tabular}
\end{table}
}
In quantitative evaluation, we score 7.57 in FID, better than SDEdit (E), InstructPix2Pix and PromptDiffusion which scores 21.67, 17.87 and 13.75.
We achieve 0.27 in CLIP DirSim,  better than baselines with  0.11/0.17/0.21 of CLIP DirSim scores. These quantitative findings show that our method generates higher-quality images with better detailed qualities and also exactly follows both language and visual editing instructions.

\noindent\textbf{Ablation study.}
We   conduct an ablation study on each of the four components of our proposed method. Namely, the Reformed Conditioned Latent Diffusion Model (RCLDM), Editing Shift Matching (ESM), Language Instruction Unification (LIU) and Selective Area Matching (SAM).
We present the qualitative results in Fig.~\ref{fig:results_ablation}. From the qualitative results, we can see that without RCLDM and ESM, the model understands the visual editing instructions much weaker, especially in out-of-domain editing. Without LIU, for two language editing instructions with the same meaning, the model produces two output edited images in different detail and quality. This difference in quality tends to increase in out-of-domain settings. Without SAM, the details of the human face are distorted making the model more vulnerable to producing outputs with worse detailed qualities.
We also include the qualitative ablation results on the testing dataset in Tab.~\ref{tab:ablation}. From observation, our method performs the best when incorporating all four components. Without SAM or LIU, the FID score increases, meaning those modules enhance the detail quality of the output generated. Without CLDM or ESM, the CLIP DirSim score decreases, showing that those two modules contribute to a better understanding in both language and visual level.
{\setlength{\tabcolsep}{8pt}
\renewcommand\arraystretch{1.25}
\begin{table}[tb]
  \centering
  \caption{Abalation Study results of comparison between our method with all components and without the Reformed Conditioned Latent Diffusion Model (RCLDM), Editing Shift Matching (ESM), Language Instruction Unification (LIU) and Selective Area Matching (SAM). The ablation study is conducted on the entire test dataset. }
  \label{tab:ablation}
  \begin{tabular}{l|c|c|c|c|c}
  \hline
& w/o RCLDM & w/o ESM & w/o LIU & w/o SAM & Ours \\
\hline
FID $\downarrow$ & 10.15 & 9.23 & 10.37  & 11.31 & 7.57 \\
\hline
CLIP DirSim $\uparrow$ & 0.13 & 0.15 & 0.23 & 0.19 & 0.27 \\
  \hline
  \end{tabular}
\end{table}
}

\section{Conclusion}
In this work, we propose InstructGIE, an image editing framework with enhanced generalization ability, improving performance in both visual and text aspects. We incorporate a VMamba-based module to enhance visual outputs and introduce an editing-shift matching strategy to augment in-context learning. Our selective area-matching technique addresses and rectifies corrupted details, while a language unification technique aligns language embeddings with editing semantics. Additionally, we compile a publicly available dataset for better generalization evaluation. Extensive experiments demonstrate our framework's superior in-context generation performance and robust generalization capability across unseen vision tasks, both quantative and qualitively.
\label{sec:conclusion}

\section*{Acknowledgement}
This work is partially supported by the Army Research Office/Army Research Laboratory via grant W911-NF-20-1-0167 to Northeastern University, National Science Foundation CCF-1937500, and the Fundamental Research Funds for the Central Universities, Peking University.

\bibliographystyle{splncs04}
\bibliography{main}

\begin{thebibliography}{10}
\providecommand{\url}[1]{\texttt{#1}}
\providecommand{\urlprefix}{URL }
\providecommand{\doi}[1]{https://doi.org/#1}

\bibitem{vqgan-mae}
Bar, A., Gandelsman, Y., Darrell, T., Globerson, A., Efros, A.: Visual prompting via image inpainting. Advances in Neural Information Processing Systems  \textbf{35},  25005--25017 (2022)

\bibitem{blau2022threat}
Blau, T., Ganz, R., Kawar, B., Bronstein, A., Elad, M.: Threat model-agnostic adversarial defense using diffusion models. arXiv preprint arXiv:2207.08089  (2022)

\bibitem{brooks2023instructpix2pix}
Brooks, T., Holynski, A., Efros, A.A.: Instructpix2pix: Learning to follow image editing instructions. In: Proceedings of the IEEE/CVF Conference on Computer Vision and Pattern Recognition. pp. 18392--18402 (2023)

\bibitem{gpt3}
Brown, T., Mann, B., Ryder, N., Subbiah, M., Kaplan, J.D., Dhariwal, P., Neelakantan, A., Shyam, P., Sastry, G., Askell, A., et~al.: Language models are few-shot learners. Advances in neural information processing systems  \textbf{33},  1877--1901 (2020)

\bibitem{chen2022re}
Chen, W., Hu, H., Saharia, C., Cohen, W.W.: Re-imagen: Retrieval-augmented text-to-image generator. arXiv preprint arXiv:2209.14491  (2022)

\bibitem{mask2former}
Cheng, B., Misra, I., Schwing, A.G., Kirillov, A., Girdhar, R.: Masked-attention mask transformer for universal image segmentation. In: Proceedings of the IEEE/CVF conference on computer vision and pattern recognition. pp. 1290--1299 (2022)

\bibitem{afhp}
Choi, Y., Uh, Y., Yoo, J., Ha, J.W.: Stargan v2: Diverse image synthesis for multiple domains. In: Proceedings of the IEEE/CVF conference on computer vision and pattern recognition. pp. 8188--8197 (2020)

\bibitem{chung2017lip}
Chung, J.S., Zisserman, A.: Lip reading in the wild. In: Computer Vision--ACCV 2016: 13th Asian Conference on Computer Vision, Taipei, Taiwan, November 20-24, 2016, Revised Selected Papers, Part II 13. pp. 87--103. Springer (2017)

\bibitem{dai2017scannet}
Dai, A., Chang, A.X., Savva, M., Halber, M., Funkhouser, T., Nie{\ss}ner, M.: Scannet: Richly-annotated 3d reconstructions of indoor scenes. In: Proceedings of the IEEE conference on computer vision and pattern recognition. pp. 5828--5839 (2017)

\bibitem{dhariwal2021diffusion}
Dhariwal, P., Nichol, A.: Diffusion models beat gans on image synthesis. Advances in neural information processing systems  \textbf{34},  8780--8794 (2021)

\bibitem{vit}
Dosovitskiy, A., Beyer, L., Kolesnikov, A., Weissenborn, D., Zhai, X., Unterthiner, T., Dehghani, M., Minderer, M., Heigold, G., Gelly, S., Uszkoreit, J., Houlsby, N.: An image is worth 16x16 words: Transformers for image recognition at scale. ICLR  (2021)

\bibitem{clipdirsim}
Gal, R., Patashnik, O., Maron, H., Chechik, G., Cohen-Or, D.: Stylegan-nada: Clip-guided domain adaptation of image generators. arXiv preprint arXiv:2108.00946  (2021)

\bibitem{openllama}
Geng, X., Liu, H.: Openllama: An open reproduction of llama (May 2023), \url{https://github.com/openlm-research/open_llama}

\bibitem{hertz2022prompt}
Hertz, A., Mokady, R., Tenenbaum, J., Aberman, K., Pritch, Y., Cohen-Or, D.: Prompt-to-prompt image editing with cross attention control. arXiv preprint arXiv:2208.01626  (2022)

\bibitem{prompt2prompt}
Hertz, A., Mokady, R., Tenenbaum, J., Aberman, K., Pritch, Y., Cohen-Or, D.: Prompt-to-prompt image editing with cross attention control. arXiv preprint arXiv:2208.01626  (2022)

\bibitem{ho2020denoising}
Ho, J., Jain, A., Abbeel, P.: Denoising diffusion probabilistic models. Advances in neural information processing systems  \textbf{33},  6840--6851 (2020)

\bibitem{survey}
Huang, Y., Huang, J., Liu, Y., Yan, M., Lv, J., Liu, J., Xiong, W., Zhang, H., Chen, S., Cao, L.: Diffusion model-based image editing: A survey. arXiv preprint arXiv:2402.17525  (2024)

\bibitem{sdsurvey}
Huang, Y., Huang, J., Liu, Y., Yan, M., Lv, J., Liu, J., Xiong, W., Zhang, H., Chen, S., Cao, L.: Diffusion model-based image editing: A survey. arXiv preprint arXiv:2402.17525  (2024)

\bibitem{ffhq}
Karras, T., Laine, S., Aila, T.: A style-based generator architecture for generative adversarial networks. In: Proceedings of the IEEE/CVF conference on computer vision and pattern recognition. pp. 4401--4410 (2019)

\bibitem{kawar2022denoising}
Kawar, B., Elad, M., Ermon, S., Song, J.: Denoising diffusion restoration models. Advances in Neural Information Processing Systems  \textbf{35},  23593--23606 (2022)

\bibitem{kawar2022enhancing}
Kawar, B., Ganz, R., Elad, M.: Enhancing diffusion-based image synthesis with robust classifier guidance. arXiv preprint arXiv:2208.08664  (2022)

\bibitem{kawar2023imagic}
Kawar, B., Zada, S., Lang, O., Tov, O., Chang, H., Dekel, T., Mosseri, I., Irani, M.: Imagic: Text-based real image editing with diffusion models. In: Proceedings of the IEEE/CVF Conference on Computer Vision and Pattern Recognition. pp. 6007--6017 (2023)

\bibitem{convnet}
Krizhevsky, A., Sutskever, I., Hinton, G.E.: Imagenet classification with deep convolutional neural networks. In: Pereira, F., Burges, C., Bottou, L., Weinberger, K. (eds.) Advances in Neural Information Processing Systems. vol.~25. Curran Associates, Inc. (2012)

\bibitem{kwon2022diffusion}
Kwon, G., Ye, J.C.: Diffusion-based image translation using disentangled style and content representation. arXiv preprint arXiv:2209.15264  (2022)

\bibitem{liu2023more}
Liu, X., Park, D.H., Azadi, S., Zhang, G., Chopikyan, A., Hu, Y., Shi, H., Rohrbach, A., Darrell, T.: More control for free! image synthesis with semantic diffusion guidance. In: Proceedings of the IEEE/CVF Winter Conference on Applications of Computer Vision. pp. 289--299 (2023)

\bibitem{vmamba}
Liu, Y., Tian, Y., Zhao, Y., Yu, H., Xie, L., Wang, Y., Ye, Q., Liu, Y.: Vmamba: Visual state space model. arXiv preprint arXiv:2401.10166  (2024)

\bibitem{liu2015faceattributes}
Liu, Z., Luo, P., Wang, X., Tang, X.: Deep learning face attributes in the wild. In: Proceedings of International Conference on Computer Vision (ICCV) (December 2015)

\bibitem{meng2021sdedit}
Meng, C., He, Y., Song, Y., Song, J., Wu, J., Zhu, J.Y., Ermon, S.: Sdedit: Guided image synthesis and editing with stochastic differential equations. arXiv preprint arXiv:2108.01073  (2021)

\bibitem{mohammad2018wikiart}
Mohammad, S., Kiritchenko, S.: Wikiart emotions: An annotated dataset of emotions evoked by art. In: Proceedings of the eleventh international conference on language resources and evaluation (LREC 2018) (2018)

\bibitem{visii}
Nguyen, T., Li, Y., Ojha, U., Lee, Y.J.: Visual instruction inversion: Image editing via visual prompting. arXiv preprint arXiv:2307.14331  (2023)

\bibitem{radford2021learning}
Radford, A., Kim, J.W., Hallacy, C., Ramesh, A., Goh, G., Agarwal, S., Sastry, G., Askell, A., Mishkin, P., Clark, J., et~al.: Learning transferable visual models from natural language supervision. In: International conference on machine learning. pp. 8748--8763. PMLR (2021)

\bibitem{clip}
Radford, A., Kim, J.W., Hallacy, C., Ramesh, A., Goh, G., Agarwal, S., Sastry, G., Askell, A., Mishkin, P., Clark, J., et~al.: Learning transferable visual models from natural language supervision. In: International conference on machine learning. pp. 8748--8763. PMLR (2021)

\bibitem{raffel2020exploring}
Raffel, C., Shazeer, N., Roberts, A., Lee, K., Narang, S., Matena, M., Zhou, Y., Li, W., Liu, P.J.: Exploring the limits of transfer learning with a unified text-to-text transformer. The Journal of Machine Learning Research  \textbf{21}(1),  5485--5551 (2020)

\bibitem{sd}
Rombach, R., Blattmann, A., Lorenz, D., Esser, P., Ommer, B.: High-resolution image synthesis with latent diffusion models. In: Proceedings of the IEEE/CVF conference on computer vision and pattern recognition. pp. 10684--10695 (2022)

\bibitem{sohl2015deep}
Sohl-Dickstein, J., Weiss, E., Maheswaranathan, N., Ganguli, S.: Deep unsupervised learning using nonequilibrium thermodynamics. In: International conference on machine learning. pp. 2256--2265. PMLR (2015)

\bibitem{song2019generative}
Song, Y., Ermon, S.: Generative modeling by estimating gradients of the data distribution. Advances in neural information processing systems  \textbf{32} (2019)

\bibitem{song2020score}
Song, Y., Sohl-Dickstein, J., Kingma, D.P., Kumar, A., Ermon, S., Poole, B.: Score-based generative modeling through stochastic differential equations. arXiv preprint arXiv:2011.13456  (2020)

\bibitem{imagebrush}
Sun, Y., Yang, Y., Peng, H., Shen, Y., Yang, Y., Hu, H., Qiu, L., Koike, H.: Imagebrush: Learning visual in-context instructions for exemplar-based image manipulation. arXiv preprint arXiv:2308.00906  (2023)

\bibitem{theis2022lossy}
Theis, L., Salimans, T., Hoffman, M.D., Mentzer, F.: Lossy compression with gaussian diffusion. arXiv preprint arXiv:2206.08889  (2022)

\bibitem{painter}
Wang, X., Wang, W., Cao, Y., Shen, C., Huang, T.: Images speak in images: A generalist painter for in-context visual learning. In: Proceedings of the IEEE/CVF Conference on Computer Vision and Pattern Recognition. pp. 6830--6839 (2023)

\bibitem{promptdiffusion}
Wang, Z., Jiang, Y., Lu, Y., He, P., Chen, W., Wang, Z., Zhou, M., et~al.: In-context learning unlocked for diffusion models. Advances in Neural Information Processing Systems  \textbf{36} (2024)

\bibitem{lsun}
Yu, F., Seff, A., Zhang, Y., Song, S., Funkhouser, T., Xiao, J.: Lsun: Construction of a large-scale image dataset using deep learning with humans in the loop. arXiv preprint arXiv:1506.03365  (2015)

\bibitem{zablotskaia2019dwnet}
Zablotskaia, P., Siarohin, A., Zhao, B., Sigal, L.: Dwnet: Dense warp-based network for pose-guided human video generation. arXiv preprint arXiv:1910.09139  (2019)

\bibitem{controlnet}
Zhang, L., Rao, A., Agrawala, M.: Adding conditional control to text-to-image diffusion models. In: Proceedings of the IEEE/CVF International Conference on Computer Vision. pp. 3836--3847 (2023)

\bibitem{zimmermann2021score}
Zimmermann, R.S., Schott, L., Song, Y., Dunn, B.A., Klindt, D.A.: Score-based generative classifiers. arXiv preprint arXiv:2110.00473  (2021)

\end{thebibliography}
\end{document}